%% file: main.tex
\documentclass{article} 
\usepackage{iclr2025_conference,times}
\usepackage{booktabs}
\usepackage{wrapfig}

\input{math_commands.tex}

\usepackage{tcolorbox}
\usepackage{hyperref}
\usepackage{url}
\usepackage{listings}
\usepackage{xcolor}
\usepackage{enumitem}
\usepackage{graphicx} 
\usepackage{wrapfig}  
\usepackage{caption}
\usepackage{subcaption}
\usepackage{booktabs}
\usepackage{multirow}
\definecolor{codegreen}{rgb}{0,0.6,0}
\definecolor{codegray}{rgb}{0.5,0.5,0.5}
\definecolor{codepurple}{rgb}{0.58,0,0.82}
\definecolor{backcolour}{rgb}{0.95,0.95,0.92}
\tcbuselibrary{skins,breakable} 

\usepackage{xcolor}

\lstdefinestyle{cleanjson}{
    basicstyle=\ttfamily\scriptsize, 
    breaklines=true,                 
    breakatwhitespace=false,         
    frame=none,                      
    backgroundcolor=\color{white},   
    showstringspaces=false,          
    aboveskip=0pt,                   
    belowskip=0pt,                   
    xleftmargin=0pt,                 
    xrightmargin=0pt,                
    columns=fullflexible             
}

\title{Efficient On-Device Agents via Adaptive \\Context Management}

\iclrfinalcopy
\usepackage{authblk}
\author{\textbf{Sanidhya Vijayvargiya}\textsuperscript{*}}
\author{\textbf{Rahul Lokesh}}
\affil{Samsung Research America}
\affil{sanidhya.v@samsung.com, rahul.lokesh@samsung.com}

\begin{document}

\maketitle
\begin{abstract}
On-device AI agents offer the potential for personalized, low-latency assistance, but their deployment is fundamentally constrained by limited memory capacity. Context in agentic settings worsens this problem due to large static tool schemas and a growing interaction history that continually expands the persistent KV cache. To maintain on-device feasibility, agents must operate near the minimum task-sufficient context, while preserving task performance. We introduce two complementary mechanisms: (1) a learned intra-session memory architecture that distills trajectories into an append-only Context State Object (CSO), preserving current state and relevant details from previous steps that may have future utility, while still supporting KV-cache reuse, and (2) a just-in-time schema-passing mechanism that loads full tool definitions only upon tool selection. We instantiate this framework by adapting 3B-parameter SLMs and profiling them on smartphone hardware. In our evaluations, CSO-based memory outperforms KV-cache compression and trained summarization while retaining performance near or even above full-history execution. Deployed on-device, our framework reduces initial tool context by over $6\times$ and interaction-context growth by $10$--$25\times$, substantially reducing initial
prompt latency and KV-cache memory, enabling persistent and capable on-device agents.
\end{abstract}

\input{1-intro}
\input{2-related-work}
\input{3-method}
\input{4-experimental-setup}
\input{5-results}
\input{6-discussion}
\input{7-conclusion}

\bibliography{iclr2025_conference}
\bibliographystyle{iclr2025_conference}
\input{appendix}

\end{document}

%% file: math_commands.tex
\usepackage{amsmath,amsfonts,bm}

\def\eqref#1{equation~\ref{#1}}

\def\1{\bm{1}}

\DeclareMathAlphabet{\mathsfit}{\encodingdefault}{\sfdefault}{m}{sl}
\SetMathAlphabet{\mathsfit}{bold}{\encodingdefault}{\sfdefault}{bx}{n}



%% file: 1-intro.tex
\section{Introduction}
\label{sec:intro}

On-device AI agents can locally access private data and applications while
providing low-latency, personalized assistance
~\citep{ferev2024device,chen2024octopusondevicelanguagemodel}. Much of the
existing on-device literature focuses on fitting capable models within device
constraints through smaller architectures, quantization, and efficient
function-calling interfaces
~\citep{xu2024ondevicelanguagemodelscomprehensive,
abdin2024phi3technicalreporthighly,erdogan2024tinyagentfunctioncallingedge}.
However, fitting the model weights is only part of the deployment problem.
Persistent agents also accumulate context from tool definitions, user messages,
tool calls, and environment observations, yet this growing context has received
comparatively less attention as a deployment bottleneck. In our 3B, 4-bit model
deployment on a Samsung Galaxy S25, retaining 10K--20K tokens requires roughly
500\,MB--1\,GB of additional KV-cache memory. Tool schemas impose a large cost
before execution begins, while interaction history creates a growing cost
throughout a task. These constraints directly conflict with two capabilities
that make agents useful: access to a broad tool set and persistent state across
long interactions.

Our motivating target is therefore the \textit{minimum task-sufficient
context}. How little context can an agent retain while still completing its
task reliably? In tool-using agents, this requires managing both the static tool schemas and dynamic interaction history.
The latter is particularly challenging because relevance is determined by
\textit{future utility}, not only by what matters to the next response. An
identifier returned by one tool may become an argument several turns later, a
failed call may constrain subsequent actions, and a completed step changes the
remaining task state. We refer to such information as
\textit{agentically relevant context}. An effective memory mechanism should
discard redundant interaction text while preserving this future-use state, along with information that tracks the agent's current progress.

Existing context-reduction approaches address parts of this problem. KV-cache
compression methods~\citep{xiao2024efficientstreaminglanguagemodels,
zhang2023h2oheavyhitteroracleefficient} retain a smaller set of
representations from the original sequence, but do not explicitly transform an
agent trajectory into a concise task state. Learned memory
and summarization~\citep{wang2025recursivelysummarizingenableslongterm,
zhou2025mem1learningsynergizememory} instead construct compact semantic
representations, but commonly maintain them through replacement-style updates.
Rewriting the persistent memory invalidates its cached prefix and may make the
temporal evolution of progress, failures, and corrections less explicit.
Separately, tool-retrieval methods such as TinyAgent
~\citep{erdogan2024tinyagentfunctioncallingedge} reduce the schemas exposed to
the model initially, but do not address the interaction history accumulated
after execution begins.

We introduce a learned intra-session memory architecture built around a
\textit{Context State Object} (CSO). After each interaction, a dedicated
LoRA adapter converts new information into a concise CSO update capturing longer-horizon state such as user goals, exact values,
completed steps, constraints, and persistent errors. A separate agentic adapter acts
on the accumulated CSO rather than the raw conversation history to complete the user task. The CSO is
strictly append-only and previous state is left unchanged while new information is
added chronologically. This provides an explicit temporal record of task state
and, importantly for on-device inference, allows previously processed CSO
tokens to remain reusable in the persistent KV cache. We address the separate
tool-context cost through schema minification and
just-in-time (JIT) schema loading, exposing concise tool descriptions initially
and loading a selected tool's full compact schema only when required.

We instantiate this framework using small language models and evaluate it
against full-history execution, KV-cache compression, and both training-free
and trained semantic summarization. We additionally test transfer on a different SLM backbone, as well as different out-of-distribution tasks from ToolAlpaca, and profile the system
on smartphone hardware. \textbf{Our contributions are:}
\vspace{-2mm}
\begin{itemize}[leftmargin=*, itemsep=0pt]

    \item \textbf{A learned intra-session memory method for tool-using agents.}
    We introduce the Context State Object (CSO), a compact task-state
    representation trained to preserve current task state and information with
    potential future utility---including exact tool outputs, constraints,
    progress, and persistent errors---while discarding redundant interaction
    history.

    \item \textbf{A cache-efficient inference architecture for persistent
    on-device agents.}
    We maintain the CSO through incremental append-only updates using separate
    State-Tracker and Executor LoRA adapters over a shared SLM. This separates
    memory maintenance from action generation, preserves the temporal evolution
    of task state, and allows previously computed persistent KV-cache prefixes
    to remain reusable across turns.

    \item \textbf{Just-in-time tool-context management.}
    We reduce the complementary static context cost through deterministic
    schema minification and a two-stage interface that exposes concise tool
    descriptions initially and loads a full compact schema only after tool
    selection.
\end{itemize}
\vspace{-3mm}

Across SLM backbones, and OOD,
these mechanisms maintain strong task performance while reducing
interaction-context growth by $10$--$25\times$ and initial tool context by over
$6\times$, with corresponding reductions in KV-cache memory and initial prompt
latency on a Samsung Galaxy S25.

%% file: 2-related-work.tex
\vspace{-3mm}
\section{Related Work}
\label{sec:related_work}
\vspace{-2mm}
\subsection{On-Device Agents and Function Calling}
\vspace{-1mm}
Small Language Models (SLMs) increasingly support function calling and
agentic execution directly on personal devices
~\citep{xu2024ondevicelanguagemodelscomprehensive,gemmateam2025gemma3technicalreport,abdin2024phi3technicalreporthighly,chen2024octopusondevicelanguagemodel}.
This line of work has primarily focused on fitting capable models within
device constraints through smaller architectures, quantization, and
specialized function-calling training. Systems such as TinyAgent
~\citep{erdogan2024tinyagentfunctioncallingedge} additionally reduce tool-interface overhead by
retrieving or exposing only a subset of relevant tools.

These approaches make local agent execution increasingly practical, but they
primarily address model size or the tool context presented before execution.
Persistent agents introduce a complementary resource cost: interaction state
accumulates across user messages, tool calls, tool observations, and delegated
responses. This cost is particularly important on-device because context
directly consumes KV-cache memory and increases inference overhead throughout
a trajectory. Our work focuses on this persistent context-management problem,
while also considering the static tool-schema cost through just-in-time schema
loading.
\vspace{-1mm}
\subsection{KV-Cache Compression}
\vspace{-1mm}
A broad class of inference methods reduces context cost by retaining a subset
of token representations or reallocating KV-cache capacity.
StreamingLLM~\citep{xiao2024efficientstreaminglanguagemodels} preserves
attention-sink and recent tokens, while H$_2$O
~\citep{zhang2023h2oheavyhitteroracleefficient} prioritizes tokens according to
accumulated attention. More recent methods such as HeadKV-R2
~\citep{fu2025headsmatterheadlevelkv} allocate cache capacity using head-specific signals
associated with retrieval and reasoning.

These methods operate directly over representations derived from the original
token sequence. This is attractive because compression can be applied without
constructing a separate semantic memory, and some methods explicitly model
which parts of the sequence are important for later reasoning. Our setting,
however, places a different requirement on context: later tool execution may
depend on sparse semantic state such as exact identifiers, completed steps,
constraints, or persistent failures. Rather than selecting which original
representations to retain, our State-Tracker constructs an explicit task-state
representation containing information expected to affect future execution.
We therefore view KV-cache compression and semantic task-state construction as
distinct approaches to reducing persistent context, and use HeadKV-R2 as our
primary token-level comparison.
\vspace{-1mm}
\subsection{Semantic Memory and Summarization}
\vspace{-1mm}
Semantic-memory systems reduce dependence on complete interaction histories by
compressing, externalizing, or reorganizing prior context. Recursive
summarization~\citep{wang2025recursivelysummarizingenableslongterm} periodically replaces accumulated
dialogue with a shorter textual summary, while systems such as
MemGPT~\citep{packer2024memgptllmsoperatingsystems} manage information across external memory tiers.
Learned approaches such as MEM1~\citep{zhou2025mem1learningsynergizememory} further show that models
can be trained to maintain compact memory representations for long-running
interactions.

Our work shares the goal of replacing verbose history with semantic memory,
but targets a different combination of requirements. First, the retained state
is trained specifically around information with potential future utility for
tool execution, rather than general conversational recall. Second, the CSO is
maintained incrementally: previously committed state is not regenerated when
new information arrives. This preserves the temporal evolution of progress,
errors, and corrections while keeping the corresponding persistent KV-cache
prefix reusable across turns. Memory maintenance is also separated from action
generation using swappable LoRA adapters over a shared base model
~\citep{hu2021loralowrankadaptationlarge}.

To distinguish these factors from the general benefit of learned semantic
compression, we directly compare against a trained replacement-summary
baseline using the same base model, training data, and teacher supervision.
This provides a closely matched comparison between regenerating complete
compressed memory and maintaining task state through append-only updates.

%% file: 3-method.tex
\vspace{-2mm}
\section{Framework for Context-Efficient On-Device Agents}
\label{sec:methodology}

\vspace{-1mm}
\subsection{System Architecture}
\label{subsec:sys_arch}
\vspace{-1mm}
We instantiate our framework for context condensation in a hybrid on-device agent powered by a Small
Language Model (SLM). Given a request, the agent can respond directly to a conversational input, invoke an
on-device tool to manipulate local data, or delegate to a cloud agent for complex tasks and reasoning. Local tools can produce exact values needed in later
steps, while cloud responses can introduce large amounts of text. Our
context-management mechanisms are independent of the routing policy and focus
on retaining the task state and tool information required for subsequent
execution.

\vspace{-2mm}
\subsection{Intra-Session Memory via Context State Object}
\label{subsec:textual_memory}
\vspace{-2mm}
Rather than retaining the raw trajectory, we maintain a structured textual
CSO containing concise longer-horizon task
state, such as user goals, exact entity values, completed steps, active constraints,
tool outcomes, and persistent errors. Information not expected to affect later
execution is omitted.
\vspace{-3mm}
\paragraph{Dual-Adapter Architecture.}
The CSO is maintained using two LoRA adapters
~\citep{hu2021loralowrankadaptationlarge} over a shared base model:
\vspace{-2mm}
\begin{itemize}[leftmargin=*, itemsep=0pt]
    \item \textbf{Executor (\texttt{LoRA\textsubscript{Exec}}):}
    generates responses, tool calls, and cloud-delegation decisions conditioned
    on the current CSO.

    \item \textbf{State-Tracker (\texttt{LoRA\textsubscript{Mem}}):}
    a 119M-parameter adapter that converts each completed interaction into a
    concise update to persistent task state.
\end{itemize}
\vspace{-2mm}
At step $t$, the Executor acts from $CSO_{t-1}$ and the current observation.
The State-Tracker then receives the new interaction and generates
only the information to retain, $\Delta_t$, appended to the state:
\[
    CSO_t = CSO_{t-1} \oplus \Delta_t .
\]
\vspace{-8mm}
\paragraph{Learning Future-Use Task State.}
The State-Tracker is trained from teacher-generated CSO updates to preserve
information that may be useful in later turns rather than uniformly summarizing
the interaction. The teacher is instructed to retain longer-horizon information
that may affect subsequent execution and is additionally provided access to the full trajectory to identify these details.
\begin{figure*}[t]
    \centering
    \begin{subfigure}[t]{0.48\textwidth}
        \vspace{0pt}
        \centering
        \vspace{-15mm}
        \includegraphics[width=\textwidth]{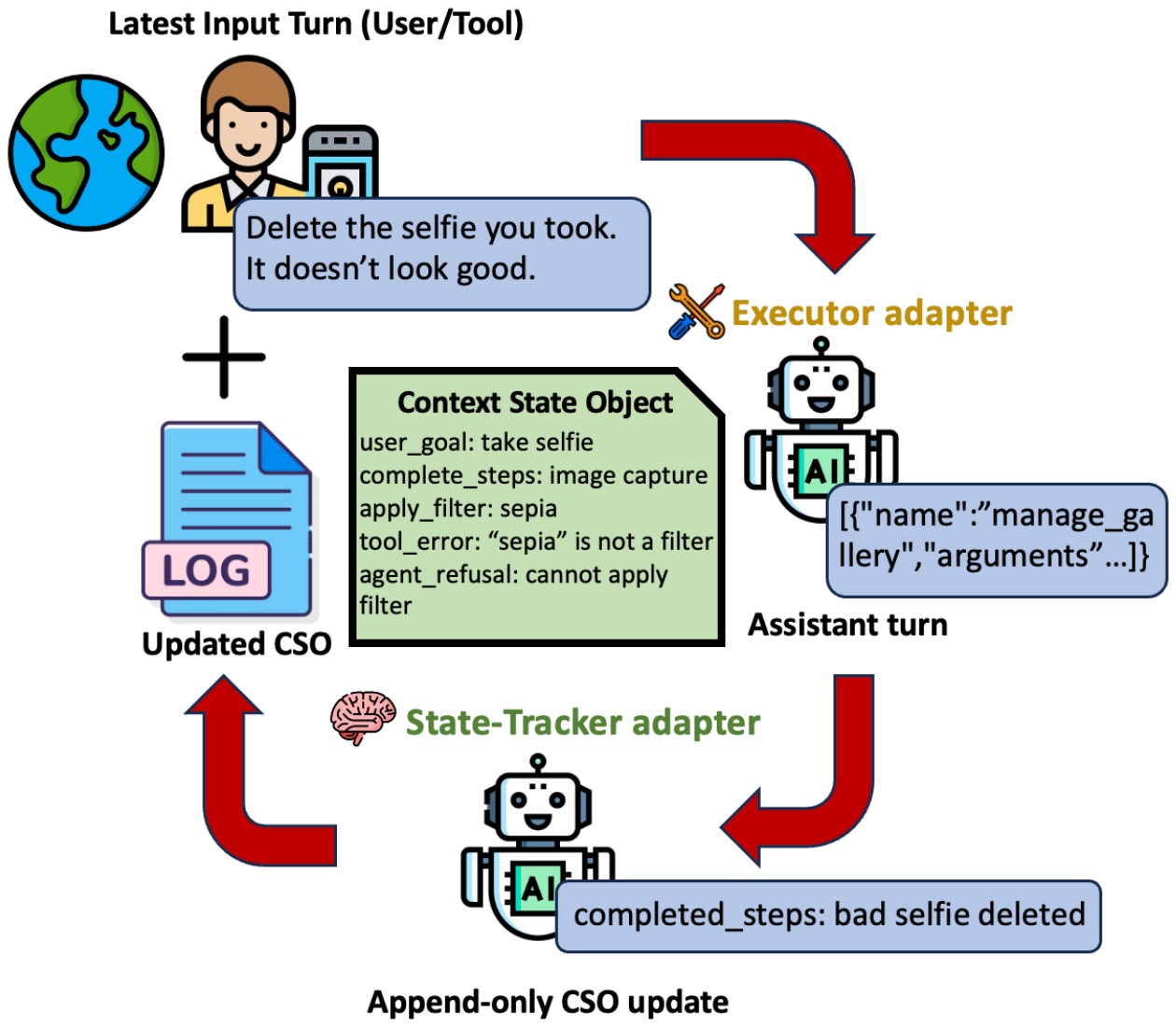}
        \vspace{-18mm}
        \caption{Interaction-history compression with the CSO.}
        \label{fig:state_management}
    \end{subfigure}
    \hfill
    \begin{subfigure}[t]{0.48\textwidth}
        \vspace{0pt}
        \centering
        \vspace{-15mm}
        \includegraphics[width=\textwidth]{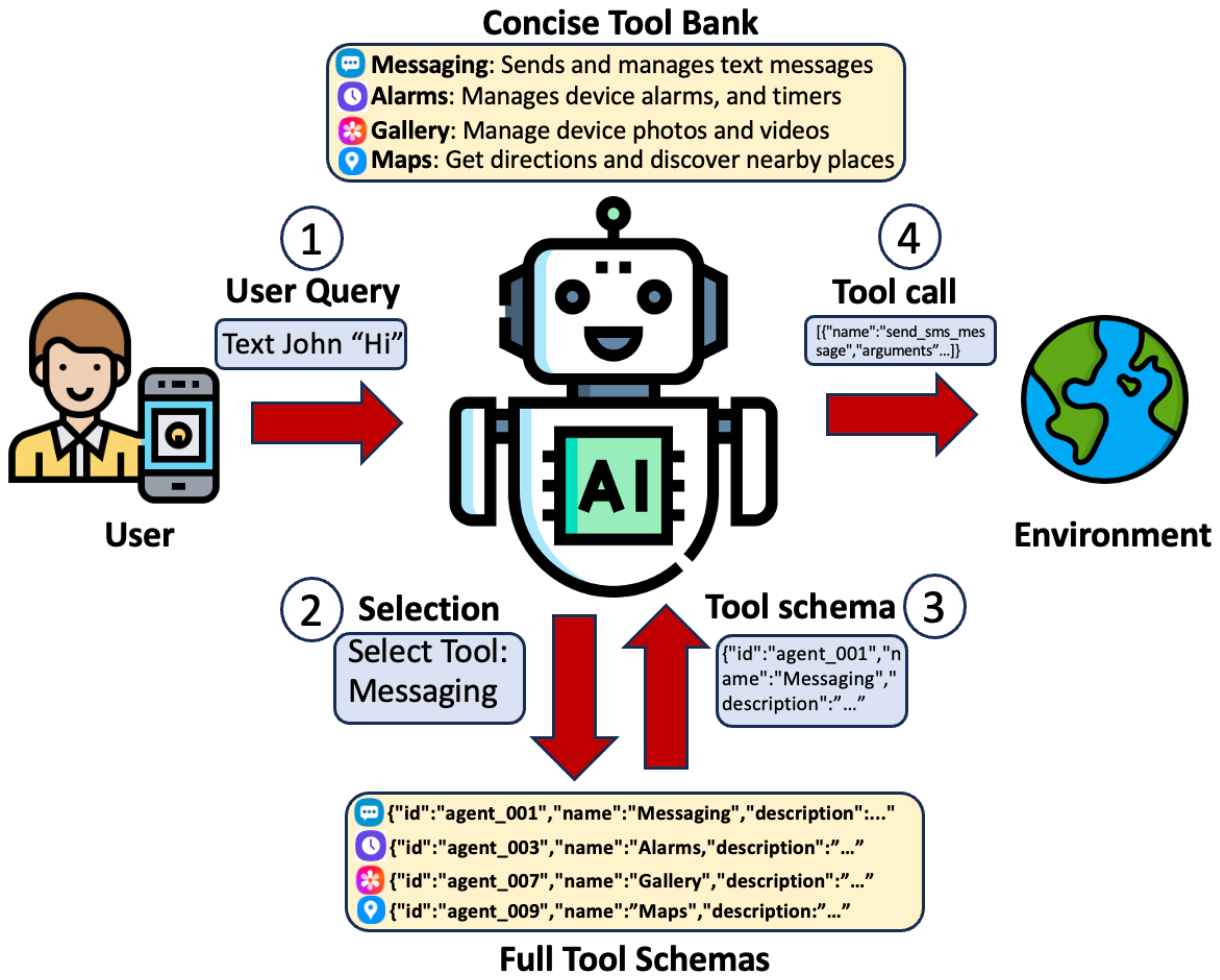}
        \vspace{-18mm}
        \caption{Just-in-time loading from a concise tool bank.}
        \label{fig:tool_selection}
    \end{subfigure}
\vspace{-2mm}
    \caption{
    Context-management mechanisms.
    (a) The State-Tracker converts interactions into compact persistent task
    state. (b) The tool interface loads a complete schema only after selection.
    }
    \label{fig:optimization_techniques}
    \vspace{-7mm}
\end{figure*}

\vspace{-5mm}
\paragraph{Append-Only State and KV-Cache Reuse.}
Unlike replacement-style memory, the CSO never rewrites previously committed
state. New updates are appended
chronologically, preserving how the task evolves while avoiding repeated
compression of previously retained information. This structure also supports incremental inference. Our runtime maintains
separate persistent KV caches for the Executor and State-Tracker. Previously
committed CSO tokens remain unchanged, so their cached representations remain
valid across turns. Current user, assistant, and tool messages are processed as
ephemeral context, after which the State-Tracker promotes the information that
should persist into $\Delta_t$. The system therefore processes newly appended
state and the current interaction without reprocessing the previously committed
CSO. The CSO uses a concise textual key-value representation and remains directly
inspectable, allowing retained information and memory failures to be analyzed.
Appendix~\ref{appendix:kv-cache} provides a complete cache-management walkthrough.

\vspace{-1mm}
\subsection{Tool Schema Management}
\label{subsec:tool_management}
\vspace{-1mm}
Tool definitions create a separate static context cost that is present before
execution begins and grows with the available capability set.
\vspace{-2mm}
\paragraph{Deterministic Schema Minification.}
Standard API representations such as OpenAPI
~\citep{OAI_OpenAPI_Specification_sw} contain formatting and metadata not
required by our function-calling interface. We deterministically convert each definition into
a compact representation retaining the tool name, description, parameters, and
required arguments while removing non-essential formatting and whitespace.
The agent is fine-tuned directly on this representation. An example is provided
in Table~\ref{tab:schema_comparison}.
\vspace{-4mm}
\paragraph{Just-in-Time Schema Loading.}
Even compact full schemas scale linearly with the number of available tools.
We therefore separate tool selection from argument generation
(Figure~\ref{fig:tool_selection}). During the \textbf{selection stage}, the
agent sees only tool names and concise descriptions. After selecting a tool, its
full compact schema is injected during the \textbf{execution stage} for
argument generation. This preserves awareness of the full capability set without placing every
schema in the initial prompt. 
\vspace{-2mm}
\subsection{Data Generation and Training}
\label{subsec:data_and_training}
\vspace{-1mm}
Training requires supervision for agent execution, JIT tool selection, and CSO
state updates. We construct this data using a multi-turn synthetic pipeline
adapted to our on-device tool environment.
\vspace{-2mm}
\paragraph{Trajectory Generation.}
We adapt the APIGen-MT pipeline
~\citep{prabhakar2025apigenmtagenticpipelinemultiturn} using Gemini 2.0
Flash~\citep{google2024gemini2flash} to synthesize multi-turn
trajectories. We construct 5,000 tools by applying
variations to 100 base tool families. For each scenario, separate instances of
the LLM simulate the user, assistant, and tool-execution environment, with the
environment including both successful and failed tool outcomes.
Trajectories are first validated against their intended tool sequence and then
filtered using an LLM-as-a-judge to remove inconsistent or unnatural
interactions. The resulting data spans multi-tool, mixed on-device/cloud, single-tool,
cloud-only, and conversational interactions, with multi-step tasks emphasized
to exercise persistent task state. Distribution details, generation prompts,
and filtering procedures are provided in
Appendix~\ref{appendix:generation_prompts}.
\vspace{-2mm}
\paragraph{Executor Training.}
We fine-tune xLAM-2 3B~\citep{zhang-etal-2025-xlam} using
LoRA~\citep{hu2021loralowrankadaptationlarge}. Each example contains
10--14 candidate tools, including the required tools, cloud agent, and
distractors. All training uses the compact schema representation. For JIT
training, the model first selects from concise tool descriptions and then
receives the selected full schema for argument generation.
\vspace{-2mm}
\paragraph{State-Tracker Training.}
The State-Tracker is trained separately from teacher-generated updates. For each interaction, the teacher receives the previous CSO, the newly observed
interaction, the task goal, and the complete generated trajectory, and produces only the incremental state
that should persist. Providing the complete trajectory during supervision allows the teacher to identify details that become useful only later in the trajectory. The supervision prioritizes token efficiency
and future utility, yielding targets for
\texttt{LoRA\textsubscript{Mem}} that transform verbose interactions into
append-only deltas.

%% file: 4-experimental-setup.tex
\vspace{-2mm}
\section{Experimental Setup}
\label{sec:experiments}
\vspace{-2mm}
We evaluate whether context reduction preserves task capability, transfers
across interaction and model distributions, and yields practical on-device
resource savings.
\vspace{-1mm}
\subsection{Evaluation Settings}
\label{subsec:dataset_and_models}
\vspace{-1mm}
\paragraph{Held-Out Tools and Simulators.}
Our primary evaluation contains 406 multi-turn tasks spanning multi-tool
execution, cloud delegation, mixed on-device/cloud tasks, and conversational
interactions. The tasks follow the same broad generation framework as training
but use 19 entirely held-out tool definitions. Evaluation is interactive, since
each method constructs a different context as the trajectory evolves. Training trajectories and State-Tracker supervision are generated using
Gemini 2.0 Flash~\citep{google2024gemini2flash}. We evaluate under both Gemini
2.0 Flash and Qwen3.5-122B, which is not used in the training-data generation
pipeline. The Gemini simulator is prompted to provide relevant information and clarifications readily, whereas the Qwen3.5 simulator is instructed not to volunteer information unless requested. 
\vspace{-2mm}
\paragraph{External Distribution and Backbone Transfer.}
We additionally evaluate on ToolAlpaca~\citep{tang2023toolalpacageneralizedtoollearning}, an independently constructed external tool and task distribution. The primary evaluated model backbone is
xLAM-2 3B~\citep{zhang-etal-2025-xlam}, but we analyze transfer by training Baseline,
Memory-Efficient, and Combined variants on Qwen3 1.7B.
\vspace{-1mm}
\subsection{Baselines and Variants}
\label{subsec:baselines}
\vspace{-1mm}
Our primary comparison set represents three approaches to interaction context:
\textbf{Baseline (FT)} retains the complete history,
\textbf{HeadKV-R2}~\citep{fu2025headsmatterheadlevelkv} performs token-level KV compression,
and \textbf{Trained Summarization} learns a compact replacement memory.
The trained summarization baseline uses the same base model, training data,
teacher supervision, and overall training protocol as our memory model, but is
trained to regenerate the complete compressed state rather than append a state
delta. We additionally include the pretrained xLAM-2 model,
training-free Recursive Summarization
~\citep{wang2025recursivelysummarizingenableslongterm},
StreamingLLM~\citep{xiao2024efficientstreaminglanguagemodels}, and
H$_2$O-style Attention Eviction
~\citep{zhang2023h2oheavyhitteroracleefficient} in our evaluations for broader coverage.

Our framework has three configurations:
\vspace{-2mm}
\begin{itemize}[leftmargin=*, itemsep=0pt]
    \item \textbf{Tool-Efficient:} compact schemas with JIT schema loading and
    full interaction history
    \item \textbf{Memory-Efficient:} append-only CSO with the complete compact
    tool bank 
    \item \textbf{Combined:} CSO-based memory together with JIT schema loading.
\end{itemize}
\vspace{-1mm}
Memory-Efficient isolates our interaction-memory contribution, while Combined
is the full context-reduction configuration.
\vspace{-1mm}
\subsection{Metrics}
\label{subsec:metrics}
\vspace{-2mm}
\paragraph{Task Performance.}
We compare the tools invoked during each trajectory against the required tools and report precision, recall, and F1. A true positive corresponds to
invoking a required tool, precision penalizes unnecessary or incorrect tool
choices, while recall measures coverage of required tools. We compute these
metrics per trajectory and macro-average precision, recall, and F1 across examples. This precision-recall decomposition is
particularly informative for JIT loading, where incorrect first-stage
selection can prevent access to the required schema. We additionally report a 1--5 LLM-as-judge measure
of overall interaction quality in Appendix~\ref{appendix:qualitative_results}. We use paired Wilcoxon signed-rank tests over matched evaluation instances for
the principal comparisons between our variants and the full-history baseline.

\paragraph{Context and Device Efficiency.}
We separately measure \textbf{initial tool context} and
\textbf{interaction-context growth}, corresponding respectively to the static
schema cost and the persistent history accumulated during execution. We profile
the Q4\_K\_M-quantized xLAM-2 3B model on a Samsung Galaxy S25 CPU using
\texttt{llama.cpp}, measuring KV-cache memory, initial prompt latency, and
State-Tracker update latency. Latency measurements are averaged across 10
trials. 

%% file: 5-results.tex
\vspace{-2mm}
\section{Results}
\label{sec:results}
\vspace{-2mm}

\subsection{Task Performance and Architectural Trade-offs}
\label{subsec:task_performance}
\vspace{-2mm}
We evaluate all methods interactively, allowing each context-management
strategy to construct its own state as the trajectory evolves.
Table~\ref{tab:gemini_results} reports results under the Gemini 2.0 Flash
simulator, while Table~\ref{tab:qwen35_results} reports the principal comparison
under the second Qwen3.5-122B simulator.

\begin{table*}[t]
\centering
\small
\renewcommand{\arraystretch}{0.92}
\setlength{\tabcolsep}{5pt}
\begin{tabular}{l @{\quad} ccc @{\quad} ccc @{\quad} ccc}
\toprule
&
\multicolumn{3}{c}{\textbf{Cloud Delegation}} &
\multicolumn{3}{c}{\textbf{Multi-Tool}} &
\multicolumn{3}{c}{\textbf{On-Device + Cloud}} \\
\cmidrule(lr){2-4}
\cmidrule(lr){5-7}
\cmidrule(lr){8-10}
\textbf{Model Variant}
& F1 & P & R
& F1 & P & R
& F1 & P & R \\
\midrule
xLAM-2 3B
& 0.41 & 0.53 & 0.63
& 0.87 & 0.86 & 0.91
& 0.57 & 0.65 & 0.63 \\

Baseline (FT)
& \textbf{1.00} & \textbf{1.00} & \textbf{1.00}
& 0.83 & 0.79 & 0.92
& 0.88 & 0.94 & 0.87 \\
\midrule
Recursive Summarization
& 0.98 & 0.98 & \textbf{1.00}
& 0.67 & 0.62 & 0.79
& 0.81 & 0.86 & 0.80 \\

StreamingLLM
& 0.97 & 0.96 & \textbf{1.00}
& 0.73 & 0.70 & 0.79
& 0.81 & 0.87 & 0.80 \\

Attention Eviction
& 0.99 & 0.98 & \textbf{1.00}
& 0.75 & 0.72 & 0.84
& 0.85 & 0.91 & 0.83 \\
\midrule
Tool-Efficient
& 0.93 & 0.97 & 0.95
& 0.86 & \textbf{0.92} & 0.88
& 0.73 & 0.85 & 0.74 \\

Memory-Efficient
& \textbf{1.00} & \textbf{1.00} & \textbf{1.00}
& 0.87 & 0.87 & 0.90
& 0.89 & \textbf{0.95} & 0.88 \\

\textbf{Combined}
& 0.99 & 0.98 & \textbf{1.00}
& \textbf{0.93} & \textbf{0.92} & \textbf{0.95}
& \textbf{0.94} & 0.92 & \textbf{0.92} \\
\bottomrule
\end{tabular}
\vspace{-1mm}
\caption{
Performance under the Gemini 2.0 Flash simulator. Qualitative interaction
scores and conversational-only results are reported in
Appendix~\ref{appendix:qualitative_results}.
}
\vspace{-4mm}
\label{tab:gemini_results}
\end{table*}

\begin{table*}[t]
\centering
\small
\renewcommand{\arraystretch}{0.92}
\setlength{\tabcolsep}{5pt}
\begin{tabular}{l @{\quad} ccc @{\quad} ccc @{\quad} ccc}
\toprule
&
\multicolumn{3}{c}{\textbf{Cloud Delegation}} &
\multicolumn{3}{c}{\textbf{Multi-Tool}} &
\multicolumn{3}{c}{\textbf{On-Device + Cloud}} \\
\cmidrule(lr){2-4}
\cmidrule(lr){5-7}
\cmidrule(lr){8-10}
\textbf{Method}
& F1 & P & R
& F1 & P & R
& F1 & P & R \\
\midrule
Baseline (FT)
& 0.94 & 0.97 & 0.91
& 0.51 & 0.75 & 0.39
& 0.63 & 0.87 & 0.49 \\

HeadKV-R2
& 0.95 & 0.97 & 0.93
& 0.59 & 0.81 & 0.46
& 0.60 & 0.88 & 0.46 \\

Trained Summarization
& 0.93 & 0.96 & 0.90
& 0.64 & 0.75 & 0.56
& 0.66 & 0.80 & 0.56 \\

Memory-Efficient
& \textbf{1.00} & \textbf{1.00} & \textbf{1.00}
& 0.69 & 0.74 & 0.65
& \textbf{0.73} & 0.86 & 0.63 \\

Combined
& \textbf{1.00} & \textbf{1.00} & \textbf{1.00}
& \textbf{0.73} & 0.90 & 0.61
& 0.72 & 0.92 & 0.59 \\
\bottomrule
\end{tabular}
\vspace{-1mm}
\caption{
Performance under the Qwen3.5-122B simulator against full-history execution,
HeadKV-R2, and matched trained replacement summarization.
}
\vspace{-4mm}
\label{tab:qwen35_results}
\end{table*}
\vspace{-2mm}
\paragraph{Comparison with Context-Compression Baselines.}
Under the Gemini simulator, Memory-Efficient matches or improves upon
full-history execution across the tool-using settings, while the training-free
compression baselines degrade more noticeably on the stateful Multi-Tool and
On-Device+Cloud tasks. Combined achieves the strongest overall performance in
these two settings while additionally reducing static tool context.

The Qwen3.5 evaluation provides a more challenging interaction setting and
shows a related but more informative pattern. Both HeadKV-R2 and trained
replacement summarization improve over full-history execution on Multi-Tool
tasks, suggesting that reducing accumulated history can itself be beneficial for performance.
Memory-Efficient nevertheless performs better than both compression baselines
across the stateful task categories and also improves upon full-history
execution. Taken together, these comparisons suggest that the benefit is not
only due to retaining fewer tokens, but also to how persistent task state is
represented.
\vspace{-2mm}
\paragraph{Effect of JIT Schema Loading.}
Combined generally shifts toward higher precision and lower recall relative to
Memory-Efficient under the Qwen3.5 evaluation. This is consistent with the
additional selection stage introduced by JIT. When the required tool is not
selected initially, its full schema is unavailable during argument generation. JIT therefore provides an additional reduction in static tool context while
introducing a stricter tool-selection bottleneck.

Paired Wilcoxon signed-rank tests show
significant differences between Memory-Efficient and the full-history Baseline
($p<0.03$), and between Combined and the Baseline ($p<0.01$).

\vspace{-2mm}
\subsection{Generalization and Transfer}
\label{subsec:transfer}
\vspace{-1mm}
We test transfer by varying the SLM backbone and the task distribution
under the Qwen3.5 simulator.
\vspace{-6mm}
\paragraph{Backbone Transfer.}
Table~\ref{tab:qwen3_transfer} repeats the core comparison using Qwen3 1.7B.
The trend observed with xLAM-2 largely persists. Memory-Efficient improves over
full-history on the Multi-Tool and On-Device+Cloud settings,
while Combined provides a further gain on Multi-Tool tasks. This indicates that
the CSO-based context-management is not specific to the xLAM-2
backbone.

\begin{table*}[t]
\centering
\small
\renewcommand{\arraystretch}{0.92}
\setlength{\tabcolsep}{5pt}
\begin{tabular}{l @{\quad} ccc @{\quad} ccc @{\quad} ccc}
\toprule
&
\multicolumn{3}{c}{\textbf{Cloud Delegation}} &
\multicolumn{3}{c}{\textbf{Multi-Tool}} &
\multicolumn{3}{c}{\textbf{On-Device + Cloud}} \\
\cmidrule(lr){2-4}
\cmidrule(lr){5-7}
\cmidrule(lr){8-10}
\textbf{Method}
& F1 & P & R
& F1 & P & R
& F1 & P & R \\
\midrule
Baseline (FT)
& \textbf{1.00} & \textbf{1.00} & \textbf{1.00}
& 0.63 & 0.81 & 0.52
& 0.67 & 0.89 & 0.54 \\

Memory-Efficient
& \textbf{1.00} & \textbf{1.00} & \textbf{1.00}
& 0.70 & 0.86 & 0.59
& 0.71 & 0.93 & 0.57 \\

Combined
& 0.98 & 0.99 & 0.97
& \textbf{0.76} & 0.92 & 0.65
& \textbf{0.72} & 0.95 & 0.58 \\
\bottomrule
\end{tabular}
\vspace{-1mm}
\caption{Backbone-transfer results using Qwen3 1.7B.}
\vspace{-6mm}
\label{tab:qwen3_transfer}
\end{table*}
\vspace{-2mm}
\paragraph{External Tool Distribution.}
\begin{wraptable}{r}{0.46\textwidth}
\vspace{-1.0em}
\centering
\small
\renewcommand{\arraystretch}{0.92}
\setlength{\tabcolsep}{6pt}
\begin{tabular}{lccc}
\toprule
\textbf{Method} & \textbf{F1} & \textbf{P} & \textbf{R} \\
\midrule
Baseline (FT)         & \textbf{0.69} & 0.56 & \textbf{0.90} \\
HeadKV-R2             & 0.54 & 0.40 & 0.83 \\
Trained Summarization & 0.60 & 0.52 & 0.71 \\
Memory-Efficient      & 0.67 & 0.63 & 0.72 \\
Combined              & 0.65 & \textbf{0.74} & 0.58 \\
\bottomrule
\end{tabular}
\caption{Performance on ToolAlpaca, an external tool and task distribution.}
\label{tab:toolalpaca_results}
\vspace{-1.0em}
\end{wraptable}
Table~\ref{tab:toolalpaca_results} evaluates transfer to ToolAlpaca, which consists of tasks with more parameters per tool call and more granular details (e.g. ticket number, plane number) in the trajectory. Memory-Efficient remains close to the
full-history Baseline while outperforming HeadKV-R2 and Trained Summarization.
Unlike the earlier evaluation, however, the compressed configurations exhibit
a clearer recall cost, showing that interaction compression is not lossless on
this distribution. Combined shifts further toward precision at the expense of
recall, consistent with the additional JIT selection constraint. These results
suggest that Memory-Efficient provides the stronger capability-context
operating point on the external distribution, while Combined represents a more
aggressive reduction in static tool context.

\vspace{-2mm}
\subsection{Context Growth and On-Device Efficiency}
\label{subsec:context_efficiency}
\begin{figure}[t]
\centering
\begin{minipage}{0.49\textwidth}
\centering
\includegraphics[width=\linewidth]{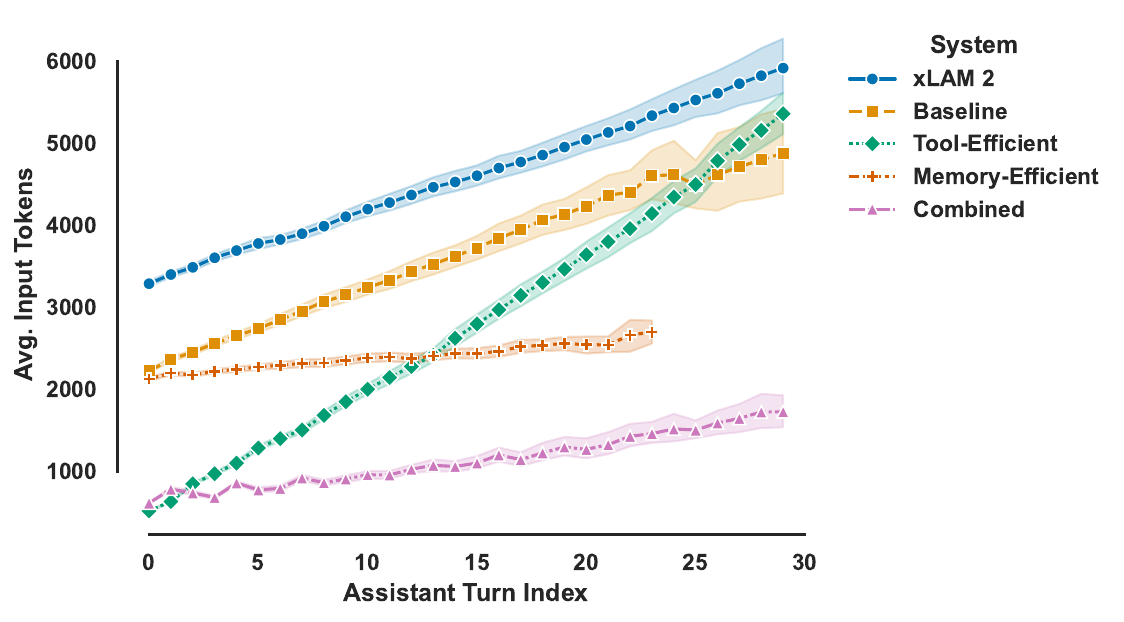}
\centerline{(a) Multi-Tool}
\end{minipage}
\hfill
\begin{minipage}{0.49\textwidth}
\centering
\includegraphics[width=\linewidth]{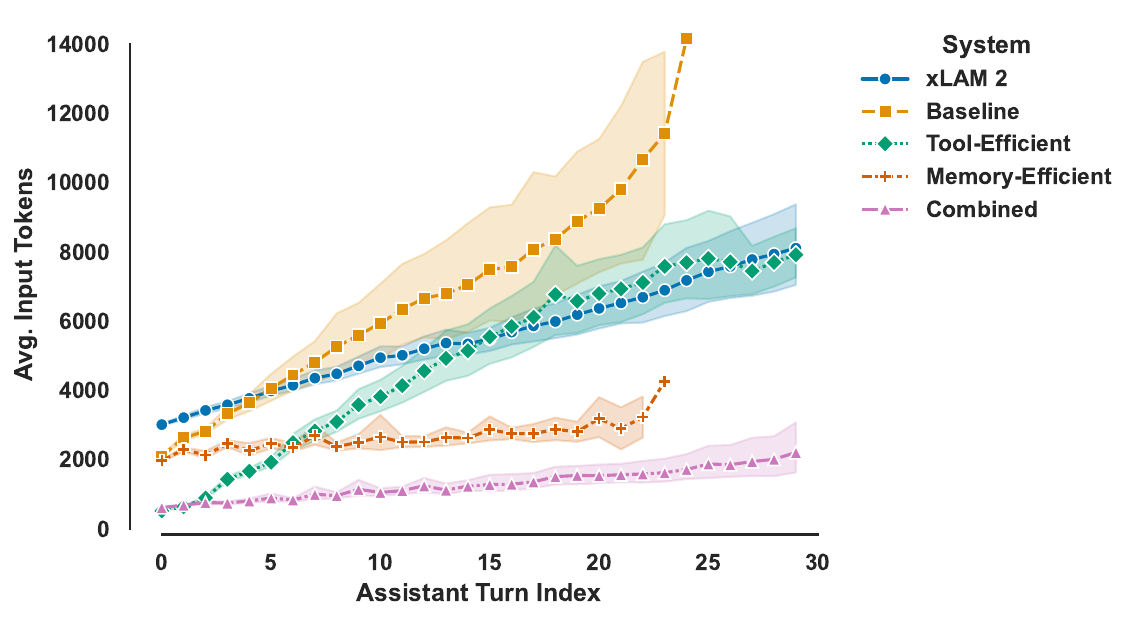}
\centerline{(b) Cloud Delegation}
\end{minipage}
\vspace{-2mm}
\caption{
Average model-visible input context over assistant turns for
(a) Multi-Tool and (b) Cloud Delegation trajectories.
Shaded regions denote 95\% confidence intervals.
}
\vspace{-5mm}
\label{fig:context_growth}
\end{figure}

\vspace{-2mm}
The two mechanisms address distinct deployment costs: JIT reduces the static
tool context at initialization, while the CSO limits the dynamic
interaction context accumulated throughout a trajectory.
\vspace{-6mm}
\paragraph{Initial Tool Context.}
In our evaluations, xLAM-2 schemas require 3,246 initial tokens on average. Deterministic schema
minification reduces this to 2,120 tokens, while JIT loading reduces the
initial tool context further to 347 tokens. JIT therefore provides a
$6.1\times$ reduction relative to compact full-schema execution and a
$9.4\times$ reduction relative to the original schema representation.
\vspace{-2mm}
\paragraph{Interaction-Context Growth.}
Figure~\ref{fig:context_growth} shows how context evolves over
multi-turn execution. In Multi-Tool scenarios, full-history execution grows to
approximately 5,000 tokens, while Memory-Efficient adds only 100--200
persistent tokens over 25 turns, corresponding to roughly a $10\times$
reduction in interaction-context growth. Tool-Efficient begins from a smaller
tool prompt but continues to accumulate raw interaction history, whereas
Combined reduces both sources of context overhead.

The difference is larger for Cloud Delegation, where verbose cloud responses
cause the full-history context to grow beyond 10,000 tokens. The State-Tracker
instead promotes selected information from these responses into concise CSO
updates. Memory-Efficient adds approximately 500 persistent tokens over these
trajectories, yielding more than a $20\times$ reduction in
interaction-context growth. Across the evaluated trajectory types, CSO-based
memory reduces interaction-context growth by approximately $10$--$25\times$.

\paragraph{On-Device Memory and Latency.}
We profile the quantized xLAM-2 3B deployment on a Samsung Galaxy S25 CPU.
The 3,246-token standard initial context requires approximately 177\,MB of
KV-cache memory, while the 347-token JIT configuration requires approximately
55\,MB and reduces initial Time-To-First-Token by approximately $5\times$.

As trajectories grow, retaining 10K--20K tokens requires roughly
500\,MB--1\,GB of KV-cache memory in our deployment. CSO-based execution
instead keeps the evaluated long-running contexts at approximately 1K--2K
active tokens, substantially reducing this growing memory cost. State tracking
adds an average 1.31\,s post-turn inference pass. Because committed CSO state
is append-only, its previously computed KV-cache prefix remains reusable on
subsequent turns.

%% file: 6-discussion.tex
\section{Discussion}
\label{sec:discussion}

Our results suggest that context efficiency for persistent agents depends not
only on retaining fewer tokens, but on preserving sparse information with
future utility, such as exact identifiers, prior failures, and completed steps.
We consider three implications for persistent-agent context management.

\paragraph{Persistent context should reflect future-use task state.}
The qualitative trajectories illustrate two different failure modes of
history-based context management. With token-level compression (e.g. KV-cache compression), we observe
cases where exact tool outputs, identifiers, or earlier error feedback needed
by a later action are no longer retained. Full-history execution avoids this
form of information loss, but must instead reason over an increasingly long
trajectory. In several such trajectories, relevant earlier information remains
present but is not effectively used by later actions, including repeated tool
calls despite preceding corrective feedback. We interpret this behavior as
consistent with attention dilution as interaction history grows.

The CSO takes a different approach by explicitly promoting selected information
from the interaction into persistent semantic task state. Rather than deciding
which representations from the original token sequence should remain
accessible, the State-Tracker determines which information should continue to
influence future execution. This distinction is reflected in the quantitative
results, where Memory-Efficient provides a stronger capability-context
operating point than the evaluated token-level compression methods across the
stateful settings. The qualitative trajectories further show how exact values,
progress, and persistent tool errors can remain explicit after the surrounding
interaction has been discarded. Appendix~\ref{appendix:error_recovery}
provides an illustrative example of this error-persistence behavior.

\paragraph{Semantic compression and incremental inference impose different requirements.}
The comparison with trained replacement summarization suggests that learned
semantic compression alone does not fully explain the performance of the CSO
architecture. The trained summarization baseline uses the same base model,
training data, teacher supervision, and overall training protocol, but
regenerates the complete compressed state after each interaction. The CSO
instead represents persistent memory as a sequence of incremental updates,
leaving previously committed state unchanged as new information is appended.

This non-destructive representation has two useful properties. First, the
evolution of the task remains explicit: completed actions, failures,
corrections, and newly observed constraints are retained in the order in which
they arise rather than being repeatedly rewritten into a replacement summary.
Our qualitative analysis suggests that this temporal structure is useful in
trajectories where later actions depend on persistent errors, corrections, or
intermediate state. This property may be particularly important in our
aggressive compression setting, where the raw interaction history is discarded
after state extraction and the CSO becomes the only persistent representation
available to subsequent execution. Repeatedly rewriting this representation can
therefore remove or blur intermediate progress, failures, and corrections that
would otherwise remain recoverable from retained history. Further, append-only state is directly compatible with
incremental inference. Because previously committed CSO tokens do not change,
their cached KV representations remain valid across turns, allowing the system
to process newly appended state without recomputing the persistent prefix. 

\paragraph{Static and dynamic context create distinct deployment trade-offs.}
Interaction memory and tool schemas contribute different forms of context
overhead. CSO-based memory limits the dynamic context accumulated as an
interaction progresses, whereas JIT schema loading reduces the static tool
context required before execution begins. The device measurements reflect this
separation. Reducing interaction history primarily lowers long-horizon
KV-cache growth, while JIT reduces initial prompt size and Time-To-First-Token.

JIT also introduces an additional tool-selection constraint. Across the
evaluations, Combined generally shifts toward higher precision and lower recall
relative to Memory-Efficient. If the required tool is not selected in the
first stage, its full schema is unavailable during argument generation, making
subsequent recovery or pivoting more difficult. We therefore view
\textit{Memory-Efficient} as the primary interaction-memory configuration,
while \textit{Combined} represents a more aggressive deployment operating
point when the static tool bank itself is also constrained. The latter is particularly helpful with a large number of tools that could not be otherwise supported, and low-latency requirement settings. This separation
allows the two mechanisms to be applied independently. CSO-based memory can
reduce growing interaction context even when the complete compact tool bank is
available, while JIT can be introduced when the cost of exposing that tool bank
justifies the additional selection bottleneck. 

%% file: 7-conclusion.tex
\section{Conclusion, Limitations, and Future Work}
\label{sec:conclusion}
\label{sec:limitations}

We introduce a context-management framework for persistent on-device agents
that represents interaction history as compact future-use task state. A learned
State-Tracker maintains this state through append-only CSO updates, allowing
semantic compression while preserving reusable KV-cache prefixes. A
complementary JIT interface reduces the static tool context required at
initialization. Across evaluations, CSO-based memory remains competitive with or exceeds full-history
execution while outperforming the evaluated context-compression baselines, and
reduces interaction-context growth by approximately $10$--$25\times$. 

\paragraph{Limitations.}
Our approach requires supervised fine-tuning and therefore depends on the
quality and coverage of State-Tracker supervision. State tracking also adds a
1.31\,s post-turn inference pass in our deployment. JIT additionally introduces a tool-selection bottleneck that can
reduce the agent's ability to pivot to new approaches. Finally, broader evaluation across
models and open-domain agent settings can further establish a wider set of use cases.

\paragraph{Future Work.}
Future work could learn memory policies directly from long-horizon task utility,
incorporate verification or uncertainty before discarding context, and develop
mechanisms for correcting, consolidating, or expiring persistent state without
sacrificing incremental KV-cache reuse.

%% file: appendix.tex
\appendix

\section{Additional Experimental Details}
\label{sec:appendix}

This appendix provides additional implementation, training, evaluation, and
qualitative-analysis details for our context-efficient agent framework. We
include representative tool definitions, model and training specifications,
evaluation protocols, baseline details, the complete State-Tracker supervision
prompt, scenario-generation prompts, a step-by-step KV-cache walkthrough,
schema-minification examples, and qualitative trajectory analyses.

\subsection{Example On-Device Tools}
\label{subsec:tool_examples}

Our on-device environment contains tools spanning device control, application
interaction, local information access, and cloud delegation. Each tool is
specified by a name, description, and structured function schema. We show a
representative subset below.

\begin{tcolorbox}[
    colback=blue!5!white,
    colframe=blue!75!black,
    title=Example: Messaging Tool,
    fonttitle=\bfseries\small,
    arc=1mm,
    boxrule=0.5pt
]
\small
\textbf{Description:} Sends and manages text messages or chats via installed
messaging applications. \par
\textbf{Function:}
\texttt{manage\_messages(action, recipient, content,
app\_preference (optional), ...)}
\end{tcolorbox}

\begin{tcolorbox}[
    colback=green!5!white,
    colframe=green!75!black,
    title=Example: Gallery \& Photo Tool,
    fonttitle=\bfseries\small,
    arc=1mm,
    boxrule=0.5pt
]
\small
\textbf{Description:} Manages photos and videos in the device gallery,
including search, viewing, sharing, and lightweight edits. \par
\textbf{Function:}
\texttt{manage\_gallery\_items(action, search\_keywords (optional),
item\_identifier (optional), ...)}
\end{tcolorbox}

\begin{tcolorbox}[
    colback=orange!5!white,
    colframe=orange!75!black,
    title=Example: Navigation Tool,
    fonttitle=\bfseries\small,
    arc=1mm,
    boxrule=0.5pt
]
\small
\textbf{Description:} Provides directions and finds nearby points of interest
using locally available map information. \par
\textbf{Function:}
\texttt{get\_local\_navigation\_info(request\_type,
destination\_or\_poi\_type, ...)}
\end{tcolorbox}

\begin{tcolorbox}[
    colback=purple!5!white,
    colframe=purple!75!black,
    title=Example: Ride Service Launcher,
    fonttitle=\bfseries\small,
    arc=1mm,
    boxrule=0.5pt
]
\small
\textbf{Description:} Launches a supported ride-service application with the
requested destination pre-filled. \par
\textbf{Function:}
\texttt{launch\_ride\_service\_app(destination, origin\_address, ...)}
\end{tcolorbox}

\begin{tcolorbox}[
    colback=red!5!white,
    colframe=red!75!black,
    title=Example: Cloud Delegation Tool,
    fonttitle=\bfseries\small,
    arc=1mm,
    boxrule=0.5pt
]
\small
\textbf{Description:} Delegates requests requiring capabilities beyond the
available on-device tools, including complex reasoning, generation, or broad
external knowledge. \par
\textbf{Function:}
\texttt{process\_with\_cloud\_llm(user\_query\_context,
reason\_for\_escalation)}
\end{tcolorbox}

\subsection{Training Specifications}
\label{appendix:training_specs}

\subsubsection{Base Models and Deployment}

Our primary experiments use xLAM-2 3B~\citep{zhang-etal-2025-xlam}, selected
for its function-calling capabilities at a scale suitable for on-device
deployment. We additionally evaluate backbone transfer using Qwen3
1.7B~\citep{yang2025qwen3technicalreport}.

All parameter-efficient fine-tuning is performed using
LlamaFactory~\citep{zheng2024llamafactory}. The primary xLAM-2 experiments are
trained on a single NVIDIA A6000 GPU with 48\,GB of VRAM.

For smartphone deployment, the xLAM-2 model is quantized using the
\texttt{Q4\_K\_M} 4-bit quantization scheme in
\texttt{llama.cpp}~\citep{ggerganov2023llama}.

\subsubsection{LoRA Hyperparameters}

We use Low-Rank Adaptation (LoRA) for parameter-efficient fine-tuning. For the
xLAM-2 experiments, the same LoRA configuration is used for the Executor,
State-Tracker, Baseline, and Tool-Efficient adapters where applicable. Each
xLAM-2 adapter contains approximately 119M trainable parameters.

\begin{table}[h!]
    \centering
    \caption{LoRA hyperparameters used for the primary xLAM-2 experiments.}
    \label{tab:lora_hyperparams}
    \begin{tabular}{lc}
        \toprule
        \textbf{Hyperparameter} & \textbf{Value} \\
        \midrule
        LoRA rank & 64 \\
        LoRA alpha & 128 \\
        LoRA dropout & 0.05 \\
        Target modules & All linear layers \\
        Trainable parameters per adapter & $\sim$119M \\
        \bottomrule
    \end{tabular}
\end{table}

\subsubsection{Fine-Tuning Hyperparameters}

Models are fine-tuned for a single epoch. Depending on the training objective,
the corresponding dataset contains approximately 150K--200K examples. We use
AdamW with a cosine learning-rate schedule.

\begin{table}[h!]
    \centering
    \caption{Primary fine-tuning hyperparameters.}
    \label{tab:ft_hyperparams}
    \begin{tabular}{lc}
        \toprule
        \textbf{Hyperparameter} & \textbf{Value} \\
        \midrule
        Epochs & 1 \\
        Learning rate & $1\times10^{-4}$ \\
        Optimizer & AdamW \\
        Learning-rate scheduler & Cosine \\
        Warmup ratio & 0.1 \\
        Effective batch size & 32 \\
        Maximum sequence length & 4096 \\
        \bottomrule
    \end{tabular}
\end{table}

A complete fine-tuning run for the primary 3B model requires approximately
4.5 hours on a single A6000 GPU.

\subsubsection{Training Data Composition}

The training distribution contains five task categories designed to cover both
simple and stateful agent interactions:

\begin{itemize}[leftmargin=*, itemsep=1pt]
    \item \textbf{Multi-Tool Tasks:} 40\%
    \item \textbf{On-Device then Cloud Tasks:} 20\%
    \item \textbf{Conversational Tasks:} 10\%
    \item \textbf{Single-Tool Tasks:} 15\%
    \item \textbf{Cloud-Only Complex Tasks:} 15\%
\end{itemize}

Multi-step interactions receive the largest share because they most directly
exercise persistent task state. All agent training uses our deterministic
token-efficient tool-schema format.

\subsection{Evaluation Protocol}
\label{appendix:evaluation_protocol}

\subsubsection{Primary Evaluation Set}

Our primary evaluation consists of 406 multi-turn tasks distributed as follows:

\begin{itemize}[leftmargin=*, itemsep=1pt]
    \item \textbf{Multi-Tool Tasks:} 181
    \item \textbf{Cloud Delegation Tasks:} 75
    \item \textbf{On-Device+Cloud Tasks:} 100
    \item \textbf{Conversational Tasks:} 50
\end{itemize}

The evaluation follows the same broad scenario-generation framework as the
training data but uses 19 entirely held-out tool definitions that do not appear
during fine-tuning. We therefore characterize this setting as held-out-tool
generalization rather than a fully independent task distribution.

Each model is evaluated interactively. This is necessary because the compared
methods construct different contexts as a trajectory develops. Full-history
execution retains previous turns directly; KV-compression methods operate on
representations of the original sequence; replacement summarization maintains a
rewritten semantic state; and our Memory-Efficient model incrementally
constructs the CSO.

\subsubsection{Simulation Settings}

Gemini 2.0 Flash is used in the synthetic training-data generation pipeline and
in our original interactive evaluation. We additionally evaluate the principal
configurations using Qwen3.5-122B as a second interaction simulator.
Qwen3.5-122B is not used anywhere in the training-data generation pipeline and
therefore provides a different interaction distribution at evaluation time.

The two simulator evaluations are reported separately because simulator
behavior changes the trajectories encountered by the agent. We therefore
compare methods primarily within the same simulator rather than interpreting
absolute scores across simulator settings.

\subsubsection{ToolAlpaca}

We additionally evaluate on ToolAlpaca to test transfer beyond our generated
on-device tool distribution. ToolAlpaca provides independently constructed
tools and task scenarios, and therefore introduces changes in both API
structure and task distribution.

We adapt the ToolAlpaca interactions to the same interactive execution
interface used by our agents while preserving the benchmark's tool semantics
and expected actions. The evaluation contains airline and retail scenarios.
Unlike the primary held-out-tool setting, no ToolAlpaca tools are used in our
agent training-data generation pipeline.

\subsubsection{Backbone Transfer}

The main experiments use xLAM-2 3B. To determine whether the context-management
interface depends on this particular function-calling model, we additionally
train Baseline, Memory-Efficient, and Combined configurations using Qwen3 1.7B.
This experiment is intended as a backbone-transfer test rather than a complete
rerun of every compression baseline.

\subsection{Baseline Implementation Details}
\label{appendix:baseline_details}

We compare methods representing three broad approaches to managing interaction
context: retaining the original history, compressing representations from the
original token sequence, and replacing the history with a semantic memory.

\paragraph{Baseline (FT).}
The fine-tuned full-history baseline uses the same agent training distribution
as our proposed models and receives the complete accumulated interaction
history. It uses the deterministic compact schema representation but performs
no interaction-history compression.

\paragraph{StreamingLLM.}
StreamingLLM~\citep{xiao2024efficientstreaminglanguagemodels} restricts the
retained sequence while preserving attention-sink tokens and recent context.
We include it in the Gemini evaluation as a training-free token-retention
baseline.

\paragraph{Attention Eviction.}
Our attention-eviction baseline follows the heavy-hitter principle of
H$_2$O~\citep{zhang2023h2oheavyhitteroracleefficient}, retaining tokens with
high accumulated attention importance while evicting lower-importance
representations under a constrained cache budget.

\paragraph{HeadKV-R2.}
HeadKV-R2 is our primary stronger KV-compression comparison. It uses
head-specific signals associated with contextual retrieval and reasoning to
allocate the available KV budget. We use a KV-cache budget of 3,000 tokens in
our experiments. Unlike the CSO, HeadKV-R2 continues to operate on
representations derived from the original sequence rather than constructing an
explicit semantic task-state representation.

\paragraph{Recursive Summarization.}
Recursive Summarization periodically converts accumulated interaction history
into a compact textual summary without training a task-specific memory
mechanism. It is included in the Gemini evaluation as a training-free semantic
compression baseline.

\paragraph{Trained Summarization.}
The Trained Summarization baseline provides a closely matched learned
replacement-memory comparison. It uses the same base model, training
interactions, teacher supervision, and overall training protocol as the
State-Tracker. The difference is the memory-update target: rather than
generating only the incremental state delta $\Delta_t$, the model regenerates
the complete compressed state after each interaction, replacing the previous
memory. This comparison tests whether learned semantic compression alone can
account for the performance of the CSO-based memory architecture.

\paragraph{Our Variants.}
\textbf{Tool-Efficient} applies deterministic schema minification and JIT
schema loading while retaining raw interaction history.
\textbf{Memory-Efficient} uses the State-Tracker/Executor architecture and
append-only CSO while making the compact full tool bank available.
\textbf{Combined} applies both interaction-memory compression and JIT schema
management.

\subsection{Evaluation Metrics}
\label{appendix:metric_details}

\paragraph{Tool-Call Metrics.}
For each trajectory, generated tool use is compared against the expected tool
behavior using set-based true positives, false positives, and false negatives.
We compute precision, recall, and F1 at the trajectory level and macro-average
the resulting values over evaluation examples.

Because precision, recall, and F1 are averaged independently across examples,
the reported aggregate F1 need not equal the harmonic mean obtained by applying
the F1 formula directly to the displayed aggregate precision and recall.

For each task category, we report:

\[
P = \frac{TP}{TP+FP},
\qquad
R = \frac{TP}{TP+FN},
\qquad
F_1 = \frac{2PR}{P+R},
\]

with the corresponding per-example scores subsequently macro-averaged.

\paragraph{Context Measurements.}
For every assistant step, we record the model-visible input context. We report
initial tool-context size separately from interaction-context growth because
the two correspond to different deployment costs. JIT schema loading primarily
changes the former, while CSO compression primarily changes the latter.

\paragraph{Device Measurements.}
On-device measurements are obtained using the xLAM-2 3B model quantized with
\texttt{Q4\_K\_M} and executed using \texttt{llama.cpp} on a Samsung Galaxy
S25 CPU. We report KV-cache memory, initial prompt latency, and State-Tracker
update latency. Latency measurements are averaged over 10 trials.

\subsection{State-Tracker Supervision Prompt}
\label{appendix:prompt_details}

The following is the teacher instruction used with Gemini 2.0 Flash to generate
State-Tracker supervision. At each completed interaction, the teacher receives
the current CSO and newly observed user, assistant, and tool messages. During
supervision generation, it additionally receives the task goal and complete
generated trajectory as reference context, allowing information that becomes
useful only in later steps to be identified. The teacher outputs only the new
lines to append to the CSO. The prompt specifies this incremental update
protocol and includes demonstrations of the desired memory behavior.
\begin{tcolorbox}[
    breakable,
    colback=black!5!white,
    colframe=black!75!black,
    title=State-Tracker Generation Prompt,
    fonttitle=\bfseries,
    arc=2mm,
    boxrule=0.5pt
]
\small

You are an expert AI assistant that analyzes new conversation turns and
generates a concise, append-only log entry. Your output MUST ONLY be the new
lines to be appended to a checklist. DO NOT repeat any information already
present in the ``Full Checklist So Far''. You are responsible for maintaining a
summary of the conversation so far such that all important details are captured
concisely. We do not want to degrade performance, but we want to be token
efficient.

\textbf{Your Rules:}
\begin{itemize}[leftmargin=*, itemsep=0pt]
    \item Be Extremely Concise: Use single words for keys
    (e.g., \texttt{user\_goal}).
    \item Log Core Facts: If new, add the user's \texttt{user\_goal} or any
    \texttt{agent\_refusals}.
    \item Log Progress: If a step is completed, add it under
    \texttt{completed\_steps} but do not make it repetitive.
    \item Log persistent, unresolved tool errors concisely to prevent the agent
    from making the same mistake again.
    \item You can format the new lines to append as you want (with or without
    keys) with the overall goal that the new lines capture the most information
    in the most concise manner.
    \item Our primary goal is creating a very concise summary but preserving all
    the key details of the conversation so no other messages are required for
    reference.
    \item You can create some new fields (do NOT overdo it) in the checklist
    such that crucial information to the task is kept track of.
    \item You DO NOT need to add the summarized user or tool message, just
    longer-horizon information about the task that will help in later turns.
    \item DO NOT output the full checklist. Only output the new lines.
    \item If there are no new lines you wish to append, output this text without
    anything else: \texttt{\# NO\_UPDATE}
    \item Follow these examples perfectly:
\end{itemize}

\hrulefill

\textbf{Example 1: Generating the first checklist entries}

\textbf{Full Checklist So Far:}\\
\texttt{\# This is the start of the conversation.}

\textbf{New Messages:}\\
user: Predict the stock market prices for Google tomorrow.\\
assistant: I can't provide specific stock market predictions. However, I can
search for information related to Google's stock and market trends.

\textbf{New Lines to Append:}\\
\texttt{user\_goal: get analyst predictions for google stock}\\
\texttt{agent\_refusals: cannot predict stock prices}

\medskip
\textbf{Example 2: Appending a completed step}

\textbf{Full Checklist So Far:}\\
\texttt{user\_goal: enable power saving mode}

\textbf{New Messages:}\\
assistant:
\texttt{\{"name":"manage\_network\_settings","arguments":
\{"action":"toggle\_wifi"\}\}}\\
tool:
\texttt{\{"status":"success","action":"toggle\_wifi",
"new\_state":"disabled"\}}

\textbf{New Lines to Append:}\\
\texttt{completed\_steps: wifi disabled}

\hrulefill

\textbf{Now, perform the task for the following input:}

\texttt{[Task Goal]}\\
\textit{\{The scenario-level task goal is injected here\}}

\texttt{[Full Generated Trajectory for Reference]}\\
\textit{\{The complete generated trajectory is injected here\}}

\texttt{[Full Checklist So Far]}\\
\textit{\{The full text of the current CSO is injected here\}}

\texttt{[New Messages]}\\
\textit{\{The text of the new user, tool, and assistant turns is injected
here\}}

\texttt{[New Lines to Append]}

\end{tcolorbox}

\subsection{Example Evaluation Scenarios}
\label{appendix:scenario_examples}

We provide representative examples from the task categories used in the
primary evaluation. Each example includes the underlying goal, the initial
simulated-user utterance, and scenario-level information used to guide the
interactive evaluation.

\begin{tcolorbox}[
    breakable,
    colback=gray!5!white,
    colframe=black!75!black,
    title=Representative Evaluation Scenarios,
    fonttitle=\bfseries
]
\small

\textbf{Multi-Tool}

\textbf{Goal:} Play a specific song and then adjust the volume.\\
\textbf{Initial Utterance:} ``Play `Bohemian Rhapsody' and turn it up.''\\
\textbf{Scenario Notes:} Requires playing a song and subsequently adjusting
the volume. The model must resolve ``turn it up'' using the preceding action.

\medskip
\hrule
\medskip

\textbf{Goal:} Find a nearby coffee shop and send the location to a friend.\\
\textbf{Initial Utterance:} ``Coffee?''\\
\textbf{Scenario Notes:} Requires resolving an ambiguous request, retrieving a
location, and transferring that result into a subsequent messaging action.

\medskip
\hrule
\medskip

\textbf{On-Device + Cloud}

\textbf{Goal:} Set a timer and then obtain recipe suggestions that fit within
that time.\\
\textbf{Initial Utterance:} ``Set a timer for 20 minutes.''\\
\textbf{Scenario Notes:} The local timer action establishes a duration that
must subsequently be transferred to the cloud-processing agent.

\medskip
\hrule
\medskip

\textbf{Goal:} Identify a song playing nearby and find similar music.\\
\textbf{Initial Utterance:} ``What song is playing right now?''\\
\textbf{Scenario Notes:} An on-device audio capability identifies the current
song; the returned title is subsequently required by the cloud-processing
agent.

\medskip
\hrule
\medskip

\textbf{Cloud Delegation}

\textbf{Goal:} Solve a mathematical problem requiring capabilities outside the
local tool set.\\
\textbf{Initial Utterance:}
``Can you help me solve this differential equation:
$dy/dx + 2xy = x$?''\\
\textbf{Scenario Notes:} The request should be delegated rather than mapped to
an unrelated local tool.

\medskip
\hrule
\medskip

\textbf{Goal:} Generate a recipe from a set of ingredients.\\
\textbf{Initial Utterance:}
``I have chicken breast, broccoli, rice, soy sauce, and ginger. Can you suggest
a creative recipe I can make?''\\
\textbf{Scenario Notes:} The request requires open-ended generation rather than
a supported on-device action.

\medskip
\hrule
\medskip

\textbf{Conversational}

\textbf{Goal:} The user shares a frustrating experience and seeks a
conversational response.\\
\textbf{Initial Utterance:}
``My internet has been down all day! It's so annoying.''\\
\textbf{Scenario Notes:} Tests whether the model avoids unnecessary tool use
when the user's intent is conversational.

\medskip
\hrule
\medskip

\textbf{Goal:} Engage in an abstract discussion.\\
\textbf{Initial Utterance:}
``Do you think there's such a thing as true free will?''\\
\textbf{Scenario Notes:} Tests non-tool conversational behavior.

\end{tcolorbox}

\subsection{Qualitative Interaction Evaluation}
\label{appendix:qualitative_results}

In addition to rule-based tool-execution metrics, the Gemini evaluation includes
an LLM-as-judge score for overall interaction quality. We treat this signal as
complementary rather than primary because it depends on model-based evaluation
and combines several aspects of trajectory quality.

\begin{table}[h!]
\centering
\small
\setlength{\tabcolsep}{5pt}
\begin{tabular}{lcccc}
\toprule
\textbf{Method}
& \textbf{Cloud}
& \textbf{Multi-Tool}
& \textbf{On-Device+Cloud}
& \textbf{Conversational} \\
\midrule
xLAM-2 3B              & 2.07 & 3.59 & 2.27 & 2.18 \\
Baseline (FT)          & 3.79 & 3.17 & 3.45 & 2.87 \\
Recursive Summarization& 3.41 & 2.26 & 1.96 & 3.23 \\
StreamingLLM           & 4.00 & 2.78 & 2.02 & 3.59 \\
Attention Eviction     & 4.18 & 2.88 & 2.16 & 3.70 \\
Tool-Efficient         & 3.55 & 3.61 & 3.32 & 3.21 \\
Memory-Efficient       & 4.11 & 3.37 & 3.64 & 3.32 \\
Combined               & 2.68 & 3.32 & 3.00 & 3.80 \\
\bottomrule
\end{tabular}
\caption{
LLM-as-judge assistant-quality scores under the Gemini simulator.
Scores range from 1 to 5, with higher values indicating better judged
interaction quality.
}
\label{tab:qualitative_results}
\end{table}

\subsection{LLM-as-Judge Prompt}
\label{appendix:llm_judge_prompt}

The following prompt is used for the complementary trajectory-quality
evaluation.

\begin{tcolorbox}[
    breakable,
    colback=black!5!white,
    colframe=black!75!black,
    title=LLM-as-Judge Evaluation Prompt,
    fonttitle=\bfseries,
    arc=2mm,
    boxrule=0.5pt
]
\small

\textbf{System Prompt}

You are an expert AI evaluator. Your task is to meticulously review a generated
conversational trajectory, focusing specifically on the assistant's
performance. Your focus is on QUALITATIVE aspects of the ASSISTANT:
naturalness, logical flow, contextual appropriateness of tool calls and their
arguments, and overall goal achievement.

You will be given:
\begin{enumerate}[leftmargin=*, itemsep=0pt]
    \item The original User Intent Description \& Scenario Notes.
    \item The Expected Tool Sequence (as abstract tool IDs).
    \item The actual Generated Conversational Trajectory (as a JSON string).
    \item A list of tools the assistant was aware of during generation.
\end{enumerate}

\textbf{Evaluation Criteria for the Assistant:}
\begin{itemize}[leftmargin=*, itemsep=0pt]
    \item \textbf{Naturalness \& Realism:}
    Are the assistant's responses natural and helpful?
    \item \textbf{Contextual Appropriateness of Tool Calls:}
    Did the assistant choose the right tool at the right time?
    \item \textbf{Argument Sensibility:}
    Are argument values sensible and correctly derived from the conversation
    context?
    \item \textbf{Clarification Handling:}
    Did the assistant ask for clarification when needed before calling a tool?
    \item \textbf{Adherence to Intent:}
    Did the assistant make reasonable efforts to address the user's intent?
    \item \textbf{Information Grounding:}
    Did the assistant hallucinate information?
\end{itemize}

\textbf{Binary Classification}

\texttt{classification}: 1 if the entire trajectory is high quality and
coherent; 0 if there are significant issues.

\textbf{Assistant Quality Score}

Rate the assistant's performance on a scale from 1 to 5
(1 = Poor, 5 = Excellent).

\textbf{Output Format}

\begin{verbatim}
{
  "classification": 0 or 1,
  "assistant_quality_score": an integer from 1 to 5,
  "feedback": "Detailed feedback and reasoning."
}
\end{verbatim}

\end{tcolorbox}

\begin{tcolorbox}[
    breakable,
    colback=black!5!white,
    colframe=black!75!black,
    title=Trajectory Evaluation Input,
    fonttitle=\bfseries,
    arc=2mm,
    boxrule=0.5pt
]
\small

Please evaluate the following trajectory based on the qualitative criteria and
scoring guidelines provided in the system prompt:

\textbf{1. User Intent Description:}\\
\textit{\{The original scenario description is injected here\}}

\textbf{2. Scenario Notes:}\\
\textit{\{The scenario notes are injected here, or N/A\}}

\textbf{3. Expected Tool Sequence:}\\
\textit{\{The intended tool sequence is injected here\}}

\textbf{4. Generated Conversational Trajectory:}\\
\textit{\{The complete generated trajectory is injected here\}}

\textbf{5. Tools Available to the Assistant:}\\
\textit{\{A formatted list of available tools is injected here\}}

Now provide your evaluation in the specified JSON format.

\end{tcolorbox}

\subsection{Qualitative Analysis of Simulated Interactions}
\label{appendix:sim_user_analysis}

We qualitatively inspect generated trajectories to understand the behavior of
the simulated user and the failure modes elicited during evaluation.

\paragraph{Goal and Persona Adherence.}
The simulator generally maintains the assigned high-level goal while producing
responses consistent with the supplied user persona. For example, scenarios
specifying frustration tend to produce impatient or corrective follow-ups,
whereas exploratory personas are more likely to provide additional information
or ask follow-up questions.

\paragraph{Ambiguity and Clarification.}
Several scenarios intentionally withhold information required for immediate
execution. These interactions test whether the agent clarifies, retrieves
missing information through an appropriate tool, or incorrectly commits to an
action. Conversational scenarios similarly test whether the agent can
distinguish between an actionable request and an utterance requiring only a
natural-language response.

\paragraph{Error-Recovery Behavior.}
Longer interactions also expose repeated-action failure modes. In some
trajectories, a tool invocation fails and the agent subsequently repeats the
same or closely related invalid action despite receiving explicit feedback.
These trajectories provide a useful setting for evaluating whether compressed
persistent state preserves failure information across later steps.

\paragraph{Qualitative Failure Patterns Across Context Strategies.}
Inspection of evaluation trajectories reveals two recurring types of context
failure. In token-level compression trajectories, we observe cases where exact
tool-returned values, identifiers, or earlier tool feedback needed by a later
action are no longer available after compression. In full-history trajectories,
the relevant information remains present in the context, but we observe cases
where later actions do not make effective use of it, including repeated invalid
tool calls despite earlier corrective feedback. We interpret the latter pattern
as consistent with attention dilution as interaction history grows, but these
examples do not isolate attention dilution as a causal mechanism. The examples
below illustrate how explicitly promoting persistent error information into the
CSO can make such feedback directly available to later execution.

Overall, qualitative inspection suggests that the simulator provides useful
goal-directed interactions spanning clarification, tool use, and error
recovery. As with any LLM-based simulator, however, the resulting interaction
distribution may contain model-specific artifacts and is not a substitute for
evaluation with real users. We therefore complement this setting with a second
simulator and ToolAlpaca in the main experiments.

\subsection{Scenario-Generation Prompts}
\label{appendix:generation_prompts}

For reproducibility, we reproduce the prompts used to construct the principal
training and evaluation scenario families. The text below reflects the prompt
templates used in the data-generation pipeline.

\subsubsection{Complex Multi-Step Scenarios}

\begin{tcolorbox}[
    breakable,
    colback=black!5!white,
    colframe=black!75!black,
    title=Prompt: Complex Multi-Step Scenarios,
    fonttitle=\bfseries,
    arc=2mm,
    boxrule=0.5pt
]
\small

You are an expert task designer creating realistic, complex user scenarios for
a new AI assistant. Generate \textit{\{num\_scenarios\}} scenarios that require
a sequence of \textbf{3 to 5 different tools} to fully resolve. These scenarios
should involve planning, research, and execution.

\textbf{Available Tools (for reference):}\\
\textit{\{A manifest of up to 40 randomly sampled, family-agnostic tools is
injected here.\}}

\textbf{Cloud Agent (for complex reasoning):}\\
\texttt{\{cloud\_processing\_agent\_id\}}

For each scenario, provide:
\begin{enumerate}[leftmargin=*, itemsep=0pt]
    \item \texttt{scenario\_description}: a high-level story of the user's
    complex goal.
    \item \texttt{user\_persona}: the user's personality and context.
    \item \texttt{initial\_user\_utterance}: the exact first user message.
    \item \texttt{intended\_tool\_sequence}: 3--5 Tool IDs representing the
    ideal path.
    \item \texttt{constraints\_and\_context}: important implicit information.
    \item \texttt{scenario\_notes}: why the scenario requires a long sequence.
\end{enumerate}

\textbf{Example:}

\begin{verbatim}
{
  "scenario_description":
    "User wants to find a good time to meet their friend Alex
     for dinner this week, suggest a restaurant, book a table,
     and get directions.",
  "user_persona":
    "Social planner, trying to organize a get-together.",
  "initial_user_utterance":
    "Find a time Alex and I are both free for dinner this week,
     suggest a good Italian place, and book it for us.",
  "intended_tool_sequence":
    ["agent_007", "agent_021", "agent_015", "agent_011"],
  "constraints_and_context":
    {"contact_name": "Alex", "cuisine_preference": "Italian"}
}
\end{verbatim}

\end{tcolorbox}

\subsubsection{Constraint-Based Single-Tool Scenarios}

\begin{tcolorbox}[
    breakable,
    colback=blue!5!white,
    colframe=blue!75!black,
    title=Prompt: Constraint-Based Single-Tool Scenarios,
    fonttitle=\bfseries,
    arc=2mm,
    boxrule=0.5pt
]
\small

You are an expert task designer. Generate
\textit{\{num\_scenarios\}} scenarios where the user's request targets a
single on-device tool, but includes complex constraints, preferences, or
conditions that the AI must understand and respect.

\textbf{Available Tools:}\\
\textit{\{A manifest of available on-device tools is injected here.\}}

For each scenario, provide:
\begin{enumerate}[leftmargin=*, itemsep=0pt]
    \item \texttt{scenario\_description}
    \item \texttt{user\_persona}
    \item \texttt{initial\_user\_utterance}
    \item \texttt{intended\_tool\_sequence}: a single Tool ID
    \item \texttt{constraints\_and\_context}
    \item \texttt{scenario\_notes}
\end{enumerate}

\textbf{Example:}

\begin{verbatim}
{
  "scenario_description":
    "User wants to schedule a meeting while avoiding a
     particular day and time range.",
  "user_persona":
    "Organized project manager, direct and specific.",
  "initial_user_utterance":
    "Book a 1-hour 'Q3 Planning' meeting with Dana,
     but make sure it's not on Friday and not before 10 AM.",
  "intended_tool_sequence": ["agent_007"],
  "constraints_and_context":
    {"avoid_day": "Friday", "earliest_time": "10:00AM"}
}
\end{verbatim}

\end{tcolorbox}

\subsubsection{On-Device then Cloud Scenarios}

\begin{tcolorbox}[
    breakable,
    colback=red!5!white,
    colframe=red!75!black,
    title=Prompt: On-Device then Cloud Scenarios,
    fonttitle=\bfseries,
    arc=2mm,
    boxrule=0.5pt
]
\small

Generate \textit{\{num\_scenarios\}} user scenarios. Each scenario MUST
require a sequence of EXACTLY 2 tools:

\begin{enumerate}[leftmargin=*, itemsep=0pt]
    \item First, an ON-DEVICE tool performs an action or gathers local
    information.
    \item Second, the CLOUD PROCESSING AGENT
    (\texttt{\{cloud\_agent\_id\}}) uses the result from the first step for
    more complex processing.
\end{enumerate}

\textbf{Available On-Device Tools:}\\
\textit{\{A manifest of available on-device tools is injected here.\}}

For each scenario, provide:
\begin{itemize}[leftmargin=*, itemsep=0pt]
    \item \texttt{scenario\_description}
    \item \texttt{initial\_user\_utterance}
    \item \texttt{intended\_tool\_sequence}
    \item \texttt{scenario\_notes}
\end{itemize}

\textbf{Example:}

\begin{verbatim}
{
  "scenario_description":
    "User took a screenshot of an error and wants the
     cloud agent to explain it.",
  "initial_user_utterance":
    "Ugh, what does this error mean? *User gestures
     to a screenshot*",
  "intended_tool_sequence": ["agent_024", "agent_026"],
  "scenario_notes":
    "Screenshot content from the local tool is passed
     to the cloud agent for analysis."
}
\end{verbatim}

\end{tcolorbox}

\subsubsection{Purely Conversational Scenarios}

\begin{tcolorbox}[
    breakable,
    colback=green!5!white,
    colframe=green!75!black,
    title=Prompt: Purely Conversational Scenarios,
    fonttitle=\bfseries,
    arc=2mm,
    boxrule=0.5pt
]
\small

Generate \textit{\{num\_scenarios\}} purely conversational scenarios. These
should simulate natural, open-ended chitchat where the user's goal is social
interaction rather than task completion. Focus on seeking opinions, sharing
feelings, or friendly banter.

For each scenario, provide:
\begin{itemize}[leftmargin=*, itemsep=0pt]
    \item \texttt{scenario\_description}
    \item \texttt{user\_persona}
    \item \texttt{initial\_user\_utterance}
    \item \texttt{intended\_tool\_sequence}: \texttt{[""]} or
    \texttt{["CONVERSATIONAL\_ROUTER"]}
    \item \texttt{scenario\_notes}
\end{itemize}

\textbf{Example:}

\begin{verbatim}
{
  "scenario_description":
    "User wants to discuss movies with the assistant
     and get its opinion.",
  "user_persona":
    "Casual movie fan, looking for a discussion.",
  "initial_user_utterance":
    "Hey, in your opinion, what's a film that everyone
     should see at least once?",
  "intended_tool_sequence": [""]
}
\end{verbatim}

\end{tcolorbox}

\subsubsection{Cloud-Only Complex Scenarios}

\begin{tcolorbox}[
    breakable,
    colback=violet!5!white,
    colframe=violet!75!black,
    title=Prompt: Cloud-Only Complex Scenarios,
    fonttitle=\bfseries,
    arc=2mm,
    boxrule=0.5pt
]
\small

You are a user scenario designer. Generate user scenarios that CLEARLY require
the advanced capabilities of the CLOUD PROCESSING AGENT
(\texttt{\{cloud\_processing\_agent\_id\}}). These tasks involve complex
reasoning, math problems, coding, creative text generation, or synthesizing
broad external knowledge.

Simpler on-device tools exist but should be insufficient for these tasks.

For each scenario, provide:
\begin{itemize}[leftmargin=*, itemsep=0pt]
    \item \texttt{scenario\_description}
    \item \texttt{user\_persona}
    \item \texttt{initial\_user\_utterance}
    \item \texttt{intended\_tool\_sequence}: only the cloud-processing tool
    \item \texttt{constraints\_and\_context} (optional)
    \item \texttt{expected\_outcome} (optional)
    \item \texttt{scenario\_notes}
\end{itemize}

\textbf{Example:}

\begin{verbatim}
{
  "scenario_description":
    "User asks their phone's AI to write a Python script
     to scrape website titles from a list of URLs.",
  "user_persona":
    "Developer looking for a quick utility.",
  "initial_user_utterance":
    "Can you write a Python script to take a list of URLs
     and grab the title tag from each one?",
  "intended_tool_sequence":
    ["cloud_processing_agent"],
  "constraints_and_context":
    {"task_type": "code_generation", "language": "Python"}
}
\end{verbatim}

\end{tcolorbox}

\subsection{Dual-Adapter KV-Cache Management}
\label{appendix:kv-cache}

We now illustrate how the append-only CSO interacts with KV-cache management.
This walkthrough corresponds to the \textbf{Memory-Efficient} configuration,
where the Executor is initialized with the compact full tool bank. Combined
uses the same CSO mechanism but additionally changes which tool schemas are
visible through JIT loading.

The implementation separates model-visible state into two components:
\textit{permanent context}, consisting of the static prefix and accumulated
CSO, and \textit{ephemeral context}, consisting of the current user/tool
interaction. Separate caches are maintained for the Executor and State-Tracker.

\begin{tcolorbox}[
    enhanced,
    breakable,
    colback=black!2!white,
    colframe=black!60!white,
    fonttitle=\bfseries,
    title=Dual-Adapter KV-Cache Walkthrough,
    arc=2mm,
    boxrule=0.5pt
]
\small

\textbf{Turn 0: Initialization}

Both adapters are initialized with their respective system prompts and the
initial CSO:

\begin{verbatim}
# This is the start of the conversation.
\end{verbatim}

The Executor's permanent cache contains its system prompt, compact tool
definitions, and initial CSO. The State-Tracker cache contains its own
instruction prefix and the same current CSO. In one representative run, these
prefixes contain approximately 1710 and 206 tokens, respectively.

\medskip
\hrule
\medskip

\textbf{Turn 1: User Creates an IT Ticket}

\textbf{User:}
``My Wi-Fi is not working, please create an IT ticket.''

The user query is processed on top of the Executor's committed persistent
prefix but is not itself committed to persistent state. The Executor then
generates a tool invocation such as:

\begin{verbatim}
<tool_call>
{
  "name": "manage_it_support_ticket",
  ...
}
</tool_call>
\end{verbatim}

The conversational tokens involved in this action are ephemeral: they are
needed to produce the current action but are not automatically committed to the
persistent context.

The State-Tracker receives the new interaction and produces an incremental
update:

\begin{verbatim}
user_goal: create_it_ticket
ticket_details:
  - issue: wifi_not_working
\end{verbatim}

This update is appended to the persistent CSO. After the tool executes, suppose
it returns:

\begin{verbatim}
{
  "status": "success",
  "ticket_id": "IT7390"
}
\end{verbatim}

The corresponding identifier is subsequently retained by the State-Tracker:

\begin{verbatim}
ticket_id: IT7390
\end{verbatim}

The raw tool observation itself need not remain in the persistent conversational
history once the relevant state has been recorded.

\medskip
\hrule
\medskip

\textbf{Turn 2: User Refers to Earlier State}

\textbf{User:}
``What's the status of that ticket?''

Before processing the new turn, the implementation rewinds the caches to their
committed permanent positions, discarding ephemeral conversational tokens from
the preceding turn. The new user request is then processed on top of the
unchanged persistent prefix containing:

\begin{verbatim}
user_goal: create_it_ticket
ticket_details:
  - issue: wifi_not_working
ticket_id: IT7390
\end{verbatim}

The Executor can therefore produce:

\begin{verbatim}
<tool_call>
{
  "name": "manage_it_support_ticket",
  "arguments": {
    "action": "check_ticket_status",
    "ticket_id": "IT7390"
  }
}
</tool_call>
\end{verbatim}

The example illustrates two properties of the CSO: an exact identifier needed
by a later action remains explicitly available after the corresponding raw
interaction is discarded, while previously committed state remains unchanged
as new information is appended.

\end{tcolorbox}

\subsection{Deterministic Tool-Schema Minification}
\label{appendix:schema_management}

Schema minification is deterministic preprocessing rather than a learned
component. It removes formatting and representational overhead while retaining
the fields used by our function-calling interface, including tool names,
descriptions, parameter definitions, and required arguments. The model is
fine-tuned directly on the resulting compact representation.
Table~\ref{tab:schema_comparison} gives a representative example.

\begin{table}[h!]
    \centering
    \small
    \caption{
    Representative standard OpenAPI schema and its deterministic compact
    representation. The example decreases from 121 to 72 tokens.
    }
    \label{tab:schema_comparison}
    \begin{tabular}{p{0.45\textwidth} p{0.45\textwidth}}
        \toprule
        \textbf{Standard OpenAPI (121 tokens)}
        &
        \textbf{Compact representation (72 tokens)}
        \\
        \midrule

\begin{lstlisting}[style=cleanjson]
{
  "type": "function",
  "function": {
    "name": "set_timer",
    "description":
      "Sets a timer for a
       specified duration.",
    "parameters": {
      "type": "object",
      "properties": {
        "duration_seconds": {
          "type": "integer",
          "description":
            "The duration of the
             timer in seconds."
        },
        "timer_name": {
          "type": "string",
          "description":
            "An optional name
             for the timer."
        }
      },
      "required":
        ["duration_seconds"]
    }
  }
}
\end{lstlisting}

&

\begin{lstlisting}[style=cleanjson]
{"type":"function",
"function":{
"name":"set_timer",
"description":"Sets a timer
for a specified duration.",
"parameters":{
"type":"object",
"properties":{
"duration_seconds":{
"type":"integer",
"description":"The duration
of the timer in seconds."},
"timer_name":{
"type":"string",
"description":"An optional
name for the timer."}},
"required":
["duration_seconds"]}}}
\end{lstlisting}

\\
\bottomrule
\end{tabular}
\end{table}

JIT schema loading applies an additional reduction. At initialization, the
Executor receives only tool names and short descriptions. Once a tool is
selected, its full compact schema is introduced for argument generation. This
reduces the static initial tool context but creates an additional selection
decision, producing the precision--recall trade-off discussed in
Section~\ref{sec:discussion}.

\subsection{Qualitative Examples of Error Persistence and Recovery}
\label{appendix:error_recovery}

We next provide an illustrative example of how persistent error information is
represented and subsequently used by the CSO. The Memory-Efficient sequence below is drawn directly
from an evaluation trajectory. The Baseline sequence is a composite constructed
from a recurring failure pattern observed across baseline trajectories and is
included only to illustrate that pattern; it is not a matched quantitative
comparison.

\begin{tcolorbox}[
    breakable,
    colback=black!5!white,
    colframe=black!75!black,
    title=Error Persistence and Recovery,
    fonttitle=\bfseries
]
\scriptsize

\textbf{User Goal:}
Find all \texttt{.txt} files in
\texttt{/documents/projects} containing the word
\texttt{configuration}.

\medskip
\hrule
\medskip

\textbf{Representative Baseline Failure Pattern}

The agent selects \texttt{manage\_device\_files} but attempts to express
content search using a parameter that is not valid for the selected operation:

\begin{verbatim}
<tool_call>{
  "name":"manage_device_files",
  "arguments":{
    "operation":"search_files",
    "path_or_search_query":"/documents/projects",
    "file_type_filter_for_search":"txt",
    "new_name_or_destination_path":"configuration"
  }
}</tool_call>
\end{verbatim}

The tool returns:

\begin{verbatim}
{
  "error":
    "Invalid request. The
     'new_name_or_destination_path'
     parameter is not applicable for
     'search_files'...",
  "success": false
}
\end{verbatim}

A recurring failure pattern in baseline trajectories is to subsequently repeat
the same invalid invocation despite the preceding tool feedback or user
correction.

\medskip
\hrule
\medskip

\textbf{Observed Memory-Efficient Recovery}

In the Memory-Efficient trajectory, an invalid invocation produces the same
type of tool feedback. The State-Tracker subsequently records:

\begin{verbatim}
tool_error:
  invalid parameter
  new_name_or_destination_path
  for search_files
\end{verbatim}

After the user confirms the desired operation, the Executor generates:

\begin{verbatim}
<tool_call>{
  "name":"manage_device_files",
  "arguments":{
    "operation":"search_files",
    "path_or_search_query":
      "/documents/projects",
    "file_type_filter_for_search":"txt"
  }
}</tool_call>
\end{verbatim}

\textbf{Analysis.}
The Memory-Efficient trajectory illustrates the intended role of the CSO:
tool feedback that may matter to future decisions can be promoted into concise
persistent state rather than depending on the complete raw trajectory remaining
available. The baseline composite illustrates a qualitatively observed
repetition failure. These examples are illustrative and do not constitute a
controlled causal comparison between append-only state and full-history
execution.

\end{tcolorbox}

\subsection{KV-Cache Memory Scaling}
\label{appendix:memory_kv}

\begin{figure}[htbp]
    \centering
    \includegraphics[width=0.6\textwidth]{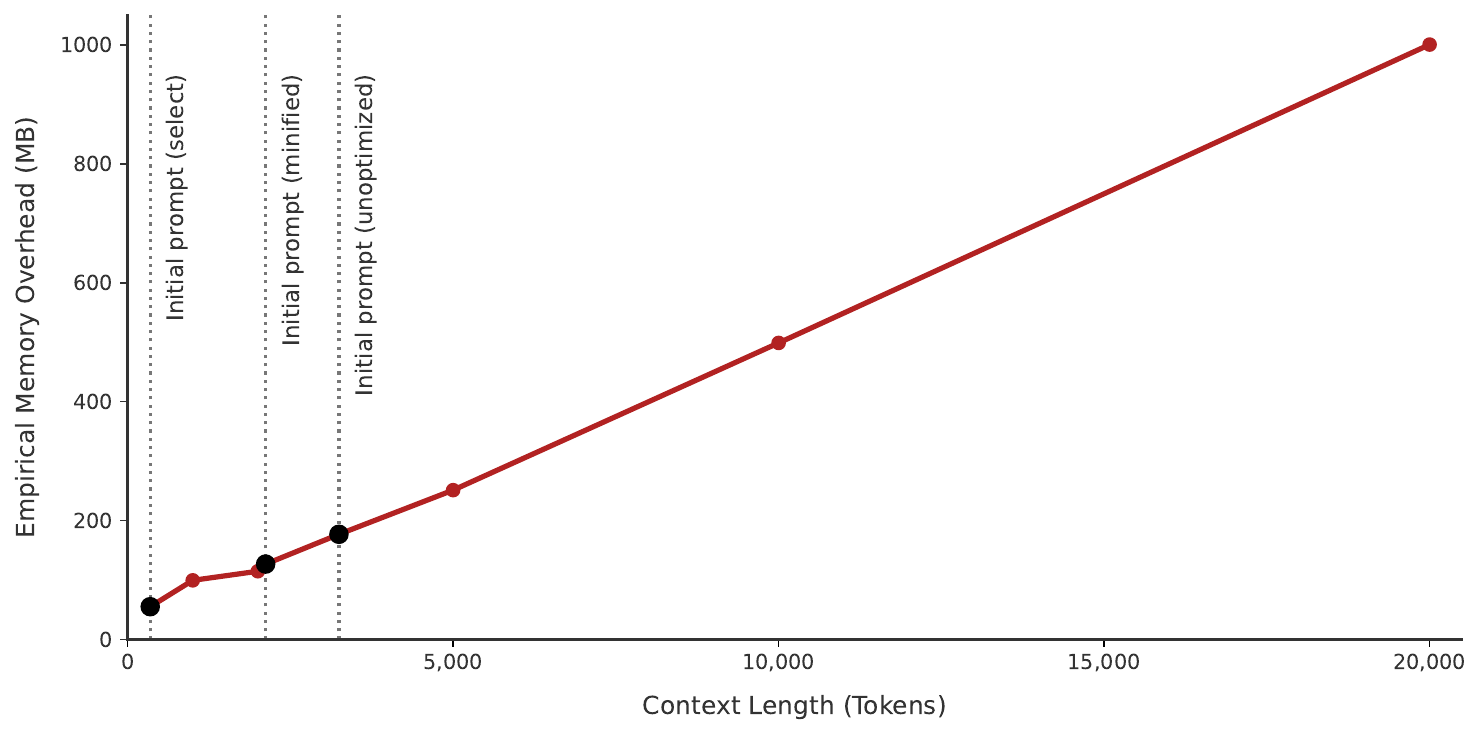}
    \caption{
KV-cache memory overhead as a function of context length for the
\texttt{Q4\_K\_M}-quantized xLAM-2 3B model under \texttt{llama.cpp}.
Retaining approximately 10K--20K tokens requires roughly 500--1,000\,MB of
KV-cache memory in our deployment. CSO-based interaction compression reduces
this long-horizon context requirement by replacing the accumulated raw
trajectory with compact persistent task state.
}
    \label{fig:memory_overhead}
\end{figure}

Figure~\ref{fig:memory_overhead} shows the measured KV-cache memory overhead as
context length increases for the quantized 3B deployment model. The
approximately linear increase makes persistent context a meaningful resource
constraint even when the underlying model nominally supports longer context
windows.

